\documentclass[11pt,letterpaper]{article}
\usepackage{acl2017}
\usepackage{times}
\usepackage{latexsym}
\usepackage{graphicx}
\usepackage{amsmath}
\usepackage{multirow}

\usepackage{subfig}
\usepackage{mdwlist}
\usepackage{enumitem}
\usepackage{amsfonts}
\usepackage{mathtools}
\usepackage{setspace}
\usepackage{stackengine}
\usepackage{array}

\aclfinalcopy

\title{A Pilot Study of Domain Adaptation Effect \\for Neural Abstractive Summarization}
\author{Xinyu Hua \and Lu Wang \\ College of Computer and Information Science \\ Northeastern University \\ Boston, MA 02115 \\ {\tt hua.x@husky.neu.edu} \quad {\tt luwang@ccs.neu.edu}}

\date{}

\begin{document}

\maketitle

\begin{abstract}
\fontsize{10}{11}\selectfont
We study the problem of domain adaptation for neural abstractive summarization. We make initial efforts in investigating what information can be transferred to a new domain. Experimental results on news stories and opinion articles indicate that neural summarization model benefits from pre-training based on extractive summaries. We also find that the combination of in-domain and out-of-domain setup yields better summaries when in-domain data is insufficient. Further analysis shows that, the model is capable to select salient content even trained on out-of-domain data, but requires in-domain data to capture the style for a target domain.
\end{abstract}

\section{Introduction}
\label{sec:intro}
Recent text summarization research moves towards producing abstractive summmaries, which better emulates human summarization process and produces more concise summaries~\cite{nenkova2011automatic}. 
Built on the success of sequence-to-sequence learning with encoder-decoder neural networks~\cite{bahdanau2014neural}, there has been growing interest in utilizing this framework for generating abstractive summaries~\cite{rush-chopra-weston:2015:EMNLP,wang-ling:2016:N16-1,takase-EtAl:2016:EMNLP2016,nallapati2016abstractive,see2017get}. 
The end-to-end learning framework circumvents efforts in feature engineering and template construction as done in previous work~\cite{ganesan2010opinosis,wang-cardie:2013:ACL2013,gerani2014abstractive,pighin-EtAl:2014:P14-1}, by directly learning to detect summary-worthy content as well as generate fluent sentences. 

Nevertheless, training such systems requires large amounts of labeled data, which creates a big hurdle for new domains where training data is scant and expensive to acquire. 
Consequently, we raise the following research questions: 

$\bullet$ {\it domain adaptation}: whether we can leverage available out-of-domain abstracts or extractive summaries to help train a neural summarization system for a new domain?

$\bullet$ {\it transferable component}: what information is transferable and what are the limitations?

\begin{figure}[t]
	\fontsize{10}{12}\selectfont
	\setlength{\tabcolsep}{0.8mm}
	\begin{tabular}{|p{75mm}|}
	\hline
	\textbf{Input (News)}:The Department of Defense has identified 441 American service members who have died since the start of the Iraq war. It confirmed the death of the following American yesterday: DAVIS, Raphael S., 24, specialist, Army National Guard; Tutwiler, Miss.; 223rd Engineer Battalion.
	\\
	
	\textbf{Abstract}: Name of American newly confirmed dead in Iraq ; 441 American service members have died since start of war.\\
	\hline
	\end{tabular}

	\begin{tabular}{|p{75mm}|}
	\hline
    \textbf{Input (Opinion)}: WHEN the 1999 United States Ryder Cup team trailed the Europeans, 10-6, going into Sunday's 12 singles matches at the Country Club outside Boston, Ben Crenshaw, the United States captain, issued a declaration of confidence in his golfers. ``I'm a big believer in faith ,'' Crenshaw said firmly in his Texas twang . `` I have a good feeling about this.'' The next day , Crenshaw' cavalry won the firsts  even singles matches. With a sudden 13-10 lead , the turnaround put unexpected pressure on the Europeans, \ldots 
    \\
	\textbf{Abstract}: Dave Anderson Sports of The Times column discusses US team's poor performance against Europe in Ryder Cup. \\
	\hline
	\end{tabular}
	\vspace{-3mm}
	\caption{\fontsize{10}{12}\selectfont A snippet of sample news story and opinion article from The New York Times Annotated Corpus~\cite{sandhausnew}. }
\label{fig:intro}
\end{figure}

In this paper, we attempt to shed some light on the above questions by investigating neural summarization on two types of documents with major difference: news stories and opinion articles from The New York Times Annotated Corpus~\cite{sandhausnew}. Sample articles and human written abstracts are shown in Figure~\ref{fig:intro}. We select a reasonably simple task on generating short news summary for multi-paragraph documents.

\noindent \textbf{Contributions.} 
We first investigate the effect of parameter initialization via pre-training on extractive summaries. A large-scale dataset consisting of 1 million article-extract pairs is collected from The New York Times for use. Experimental results show that this step improves summarization performance measured by ROUGE~\cite{lin2004rouge} and BLEU~\cite{papineni2002bleu}.
 
We then treat news stories as source domain and opinion articles as target domain, and make initial tries for understanding the feasibility of domain adaptation. Importantly, by testing on opinion article summarization, the model leveraging data from both source and target domains yields better performance than in-domain trained model when in-domain training data is rare. 
Furthermore, we interpret the learned model to understand what information is transferred to a new domain. 
In general, a model trained on out-of-domain data can learn to detect summary-worthy content, but may not match the generation style in the target domain. Concretely, we observe that the model trained on news domain pays similar amount of attention to summary-worthy content (i.e., words reused by human abstracts) when tested on news and opinion articles. 
{On the other hand, human writers tend to employ new words unseen from the input when constructing opinion abstracts. End-to-end evaluation results imply that the model trained on out-of-domain data fails to capture this aspect.}

The above observations suggest that the neural summarization model learns to 1) identify salient content, and 2) generate summaries with a style as in the training data. The first element might be transferable to a new domain, while not so much for the second.

\section{The Neural Summarization Model}
\label{sec:model}
In this work, we choose the attentional sequence-to-sequence model with pointer-generator mechanism~\cite{see2017get} for study. Briefly, the model learns to generate a sequence of tokens $\{y_i\}$ based on the following conditional probability:
$ p(y_i=w|y_1, \ldots, y_{i-1}, x) = p_{gen}P_{vocab}(w) + (1 - p_{gen})\sum_{i:w_i=w}{a_i^t} $

Here $P_{vocab}(w)$ denotes the probability to generate a new word from vocabulary, $p_{gen}$ is a learned parameter that chooses between generating and copying, depending on the hidden states and attention distribution. This model enhances the original attention model~\cite{bahdanau2014neural} by incorporating pointer-network~\cite{vinyals2015pointer}, which allows the decoder to copy accurate information from input. Due to space limitation, we refer the readers to original paper~\cite{see2017get} for model details. 

For experiments, we employ bidirectional recurrent neural network (RNN) as encoder and unidirectional RNN as decoder, both implemented by Long Short Term Memory (LSTM) with 256 hidden units. Input and output data are lowercased as described in \cite{see2017get}.

\section{Datasets and Experimental Setup}
\label{sec:setup}
\noindent \textbf{Primary Data.} Our primary data source is The New York Times Annotated Corpus~\cite{sandhausnew} (henceforth called NYT-annotated). {Compared with other commonly used dataset for abstractive summarization, NYT-annotated has more variation in its abstracts, such as paraphrase and generalization. It also comes with other human labels we could use to characterize the type of articles. The whole dataset consists of 1.8 million articles, of which 650,000 are annotated with human constructed abstracts.} 
Articles longer than 15 tokens and abstracts longer than 10 tokens are extracted for use in our study (as in Figure~\ref{fig:intro}).

\begin{figure}[t]
\hspace{-3mm}
\subfloat
{	
	\hspace{-2mm}
    \includegraphics[width=42mm,height=31mm]{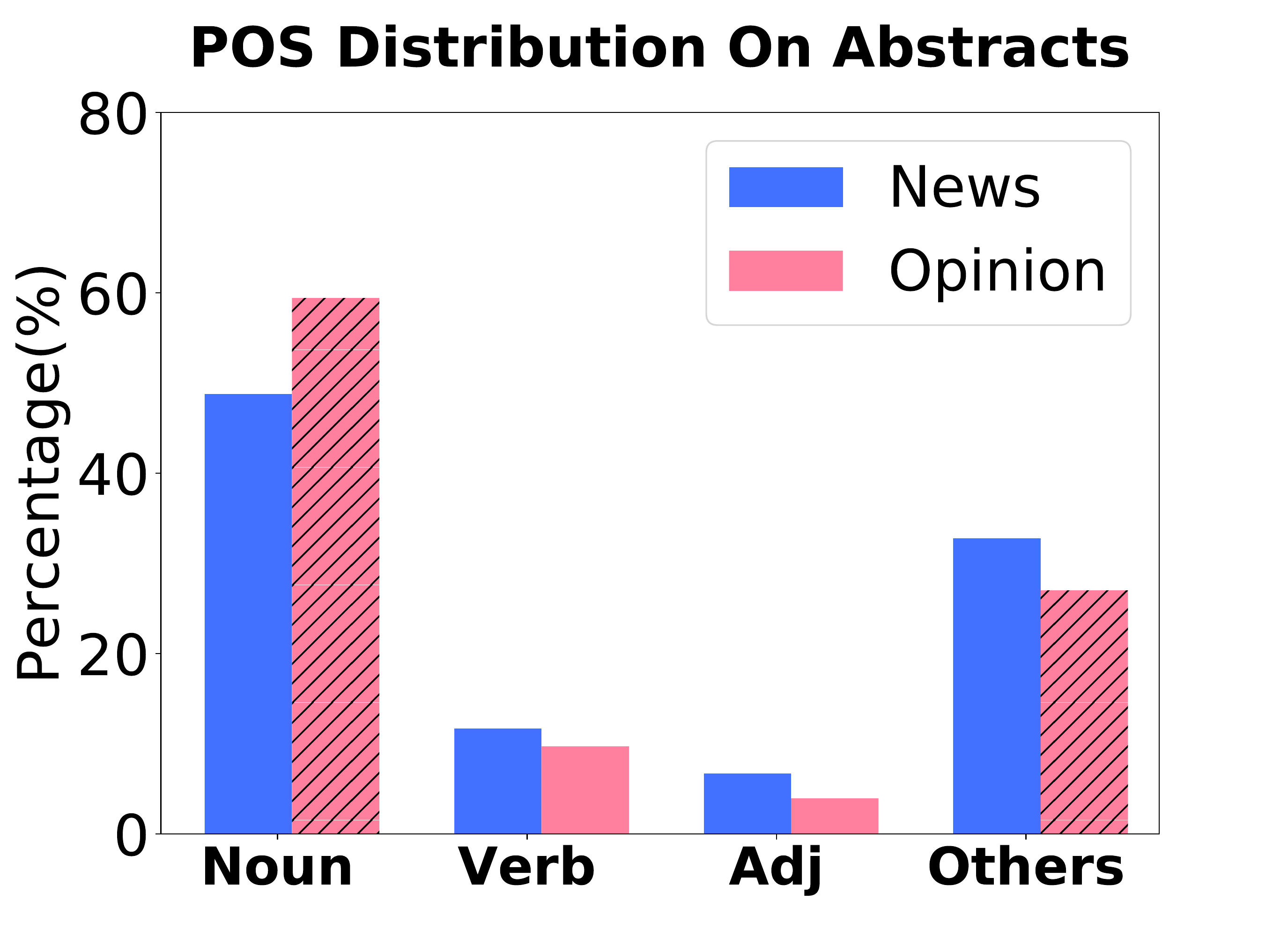}
}
\subfloat
{	
	\hspace{-3mm}
    \includegraphics[width=42mm,height=31mm]{data_5.pdf}
}
 \caption{\fontsize{10}{12}\selectfont [Left] Part-of-speech (POS) distribution for words in abstracts. [Right] Percentage of words in abstracts that are reused from input, per POS and all words. \textsc{Opinion} abstracts generally reuse less words.}
\label{fig:pos_dist}
\end{figure}
{
The resulting dataset are further separated into two types based on their taxonomy tags\footnote{The corpus comes with taxonomic classifiers tags. Articles with tag ``News" are treated as news stories; for the rest, the ones with ``Opinion'',``Editorial'', or ``Features" are treated as opinion articles.}: \textsc{News} stories and \textsc{Opinion} articles. We believe these two types of documents are different enough in terms of topics, summary style, and lexical level language use, that they could be treated as different domains for our study.}
We collected 100,824 articles for \textsc{News} which is treated as source domain, and 51,214 for \textsc{Opinion} as target domain. 
The average length for documents of \textsc{News} is 680.8 tokens, and 785.6 tokens for \textsc{Opinion}. The average lengths for abstracts are 23.14 and 19.13 for \textsc{News} and \textsc{Opinion}. 

{We also make use of the section tag, such as \emph{Business, Sports, Arts}, to calculate the topic distribution for these two domains. About 57\% of the documents of \textsc{News} are about \emph{Sports}, whereas more than 78\% documents of \textsc{Opinion} are about \emph{Arts}. We also observe different levels of subjectivity based on the percentage of strong subjective words taken from MPQA lexicon \cite{wilson2005recognizing}.  On average 4.1\% of the tokens in \textsc{Opinion} articles are strong subjective, compared to 2.9\% for \textsc{News} stories. This shows the topics and word usage are essentially different between these two domains. \vspace{1mm}
}

\begin{figure}[t]
\hspace{-3mm}
\subfloat
{	
	\hspace{-2mm}
    \includegraphics[width=42mm,height=31mm]{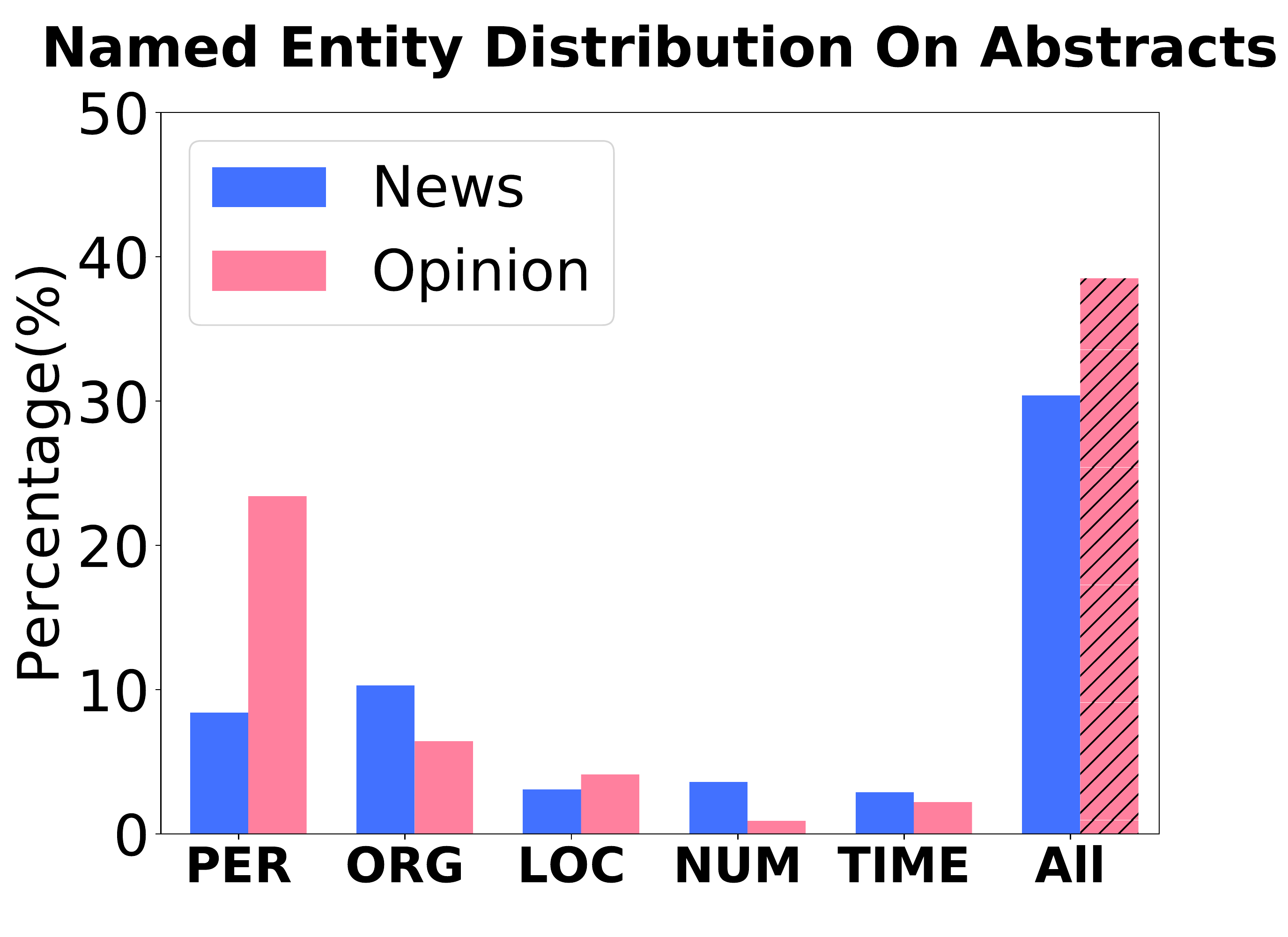}
}
\subfloat
{	
	\hspace{-3mm}
    \includegraphics[width=42mm,height=31mm]{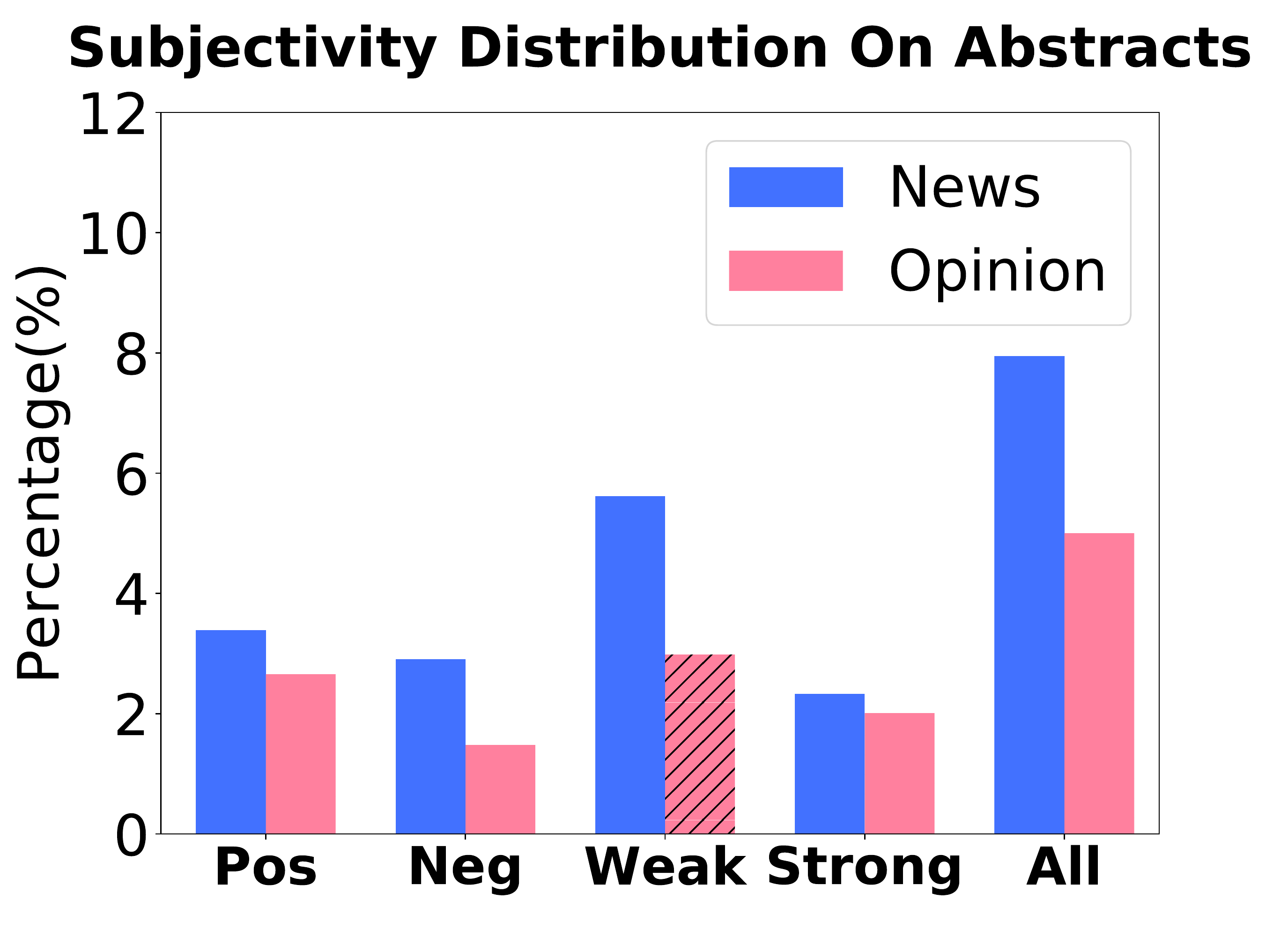}
}
 \caption{\fontsize{10}{12}\selectfont Named Entities distribution (left) and subjective words distribution (right) in abstracts. 
 More PERSON, less ORGANIZATION, and less subjective words are observed in \textsc{Opinion}.}
\label{fig:ne_subj_dist}
\end{figure}

\noindent \textit{Characterizing Two Domains.} 
Here we characterize the difference between \textsc{News} and \textsc{Opinion} by analyzing the distribution of word types in abstracts and how often human reuse words from input text to construct the summaries. Overall, 81.3\% of the words in \textsc{News} abstracts are reused from input, compared with 75.8\% for \textsc{Opinion}. 
The distribution for words of different part-of-speech is displayed on the left of Figure~\ref{fig:pos_dist}, which shows that there are relatively more Nouns in \textsc{Opinion}. In the same figure, we display the percentage of words in abstract that are reused from input, which suggests that human tends to reuse more nouns and verbs for \textsc{News} abstracts. 
Furthermore, the distribution of Named Entities words and subjective words in abstracts are depicted in Figure~\ref{fig:ne_subj_dist}.

\paragraph{Model Pre-training Dataset.} 
We further collect lead paragraphs and article descriptions for 1,435,735 articles from The New York Times \mbox{API\footnotemark.} \footnotetext{https://developer.nytimes.com}About 71\% of these descriptions are the first sentences in the lead paragraphs, and thus can be considered as extractive summaries. About one million lead paragraph and description pairs are retained for pre-training\footnote{{Unsupervised language model \cite{ramachandran2016unsupervised} can also be used for parameter initialization before our pre-training step. 
Here our goal is to allow the model to learn searching for summary-worthy content, in addition to grammaticality and language fluency.}
} (henceforth  NYT-extract). 
 
\paragraph{Training Setup.} 
We randomly divide NYT-annotated into training (75\%), validation (15\%), and test (10\%) for both news and opinion.
Experiments are conducted with the following setups: \\
1) {\textsc{In-Domain}}: Training and testing are done in the same domain, for \textsc{News} and \textsc{Opinion};
2) {\textsc{Out-Of-Domain}}: training on source domain \textsc{News}, and testing on target domain \textsc{Opinion}; 
and 3) {\textsc{Mix-Domain}}: training on source domain \textsc{News} and then on target domain \textsc{Opinion}, and testing on \textsc{Opinion}. Training stops when the trend of loss function on validation set starts increasing.

\paragraph{Evaluation Metrics.} 
We use automatic evaluation on recall-oriented ROUGE~\cite{lin2004rouge} and precision-oriented BLEU~\cite{papineni2002bleu}. We consider ROUGE-2 which measures bigram recall, and ROUGE-L which takes into account the longest common subsequence. We also evaluate on BLEU which measures precision up to bigrams.

\section{Results}
\label{sec:result}
\noindent \textbf{Effect of Pre-training with Extracts.} 
We first evaluate whether pre-training can improve summarization performance for \textsc{In-Domain} setups, where we initialize model parameters by training on NYT-extract for about 20,000 iterations. Otherwise, parameters are randomly initialized. 
Results are displayed in Table~\ref{tab:indomain}. We also consider two baselines, \textsc{Baseline1} outputs the first sentence, \textsc{Basline2} selects the first 22 (news) and 15 (opinion) tokens (with similar lengths as human summaries).
As can be seen, the pre-training step improves performance for \textsc{news}, whereas the performance on \textsc{opinion} remains roughly the same. This might be due to the fact that news abstracts reuse more words from input, which are closer to extractive summaries than opinion abstracts. 

\begin{table}[ht]
{
\fontsize{10}{12}\selectfont
\setlength{\tabcolsep}{1.3mm}

\begin{tabular}{|p{28mm}|l|l|l|l|}
    \hline
    & R-2 & R-L & BLEU & Avg Len\\
    \hline
    \textit{Test on News} & & & & \\
    \textsc{Baseline1} & 23.5 & {\bf 35.4} & 19.9 & 28.94 \\
    \textsc{Baseline2} &  19.5 & 30.1 & 19.5 & 22.00 \\
    \textsc{In-Domain} &  23.3 & 34.1 & 21.3 & 22.08 \\
    \textsc{In-Domain} + pre-train & {\bf 24.2} & 34.5 & {\bf 22.4} & 21.59\\
    \hline
    \hline
    \textit{Test on Opinion} & & & & \\
    \textsc{Baseline1} & 17.9 & 26.6 & 11.4 & 28.18 \\
    \textsc{Baseline2} & 12.9 & 20.5 & 11.7 & 15.00 \\
    \textsc{In-Domain} & 19.8 & {\bf 31.9} & {\bf 19.9} & 14.60 \\
    \textsc{In-Domain} + pre-train & {\bf 19.9} & 31.8 & 19.4 & 14.22\\
    \hline
\end{tabular}

\caption{\fontsize{10}{12}\selectfont Evaluation based on ROUGE-2 (R-2), ROUGE-L (R-L), and BLEU (multiplied by 100) for in-domain training. 
}
\label{tab:indomain}
}
\end{table}

\begin{figure}[t]
\hspace{-3mm}
\subfloat
{	
    \includegraphics[width=42mm,height=38mm]{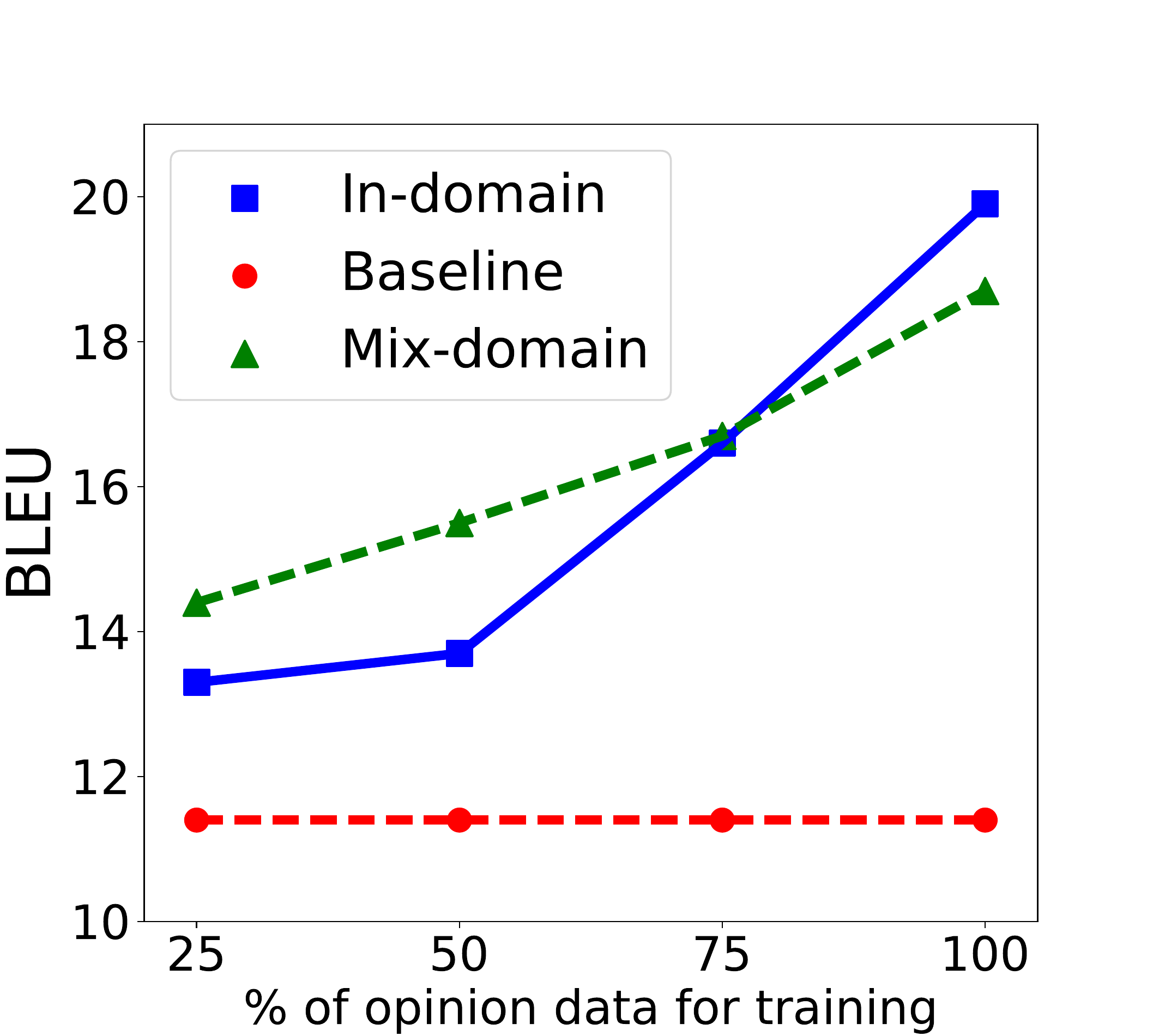}
}
\subfloat
{	
	\hspace{-5mm}
    \includegraphics[width=42mm,height=38mm]{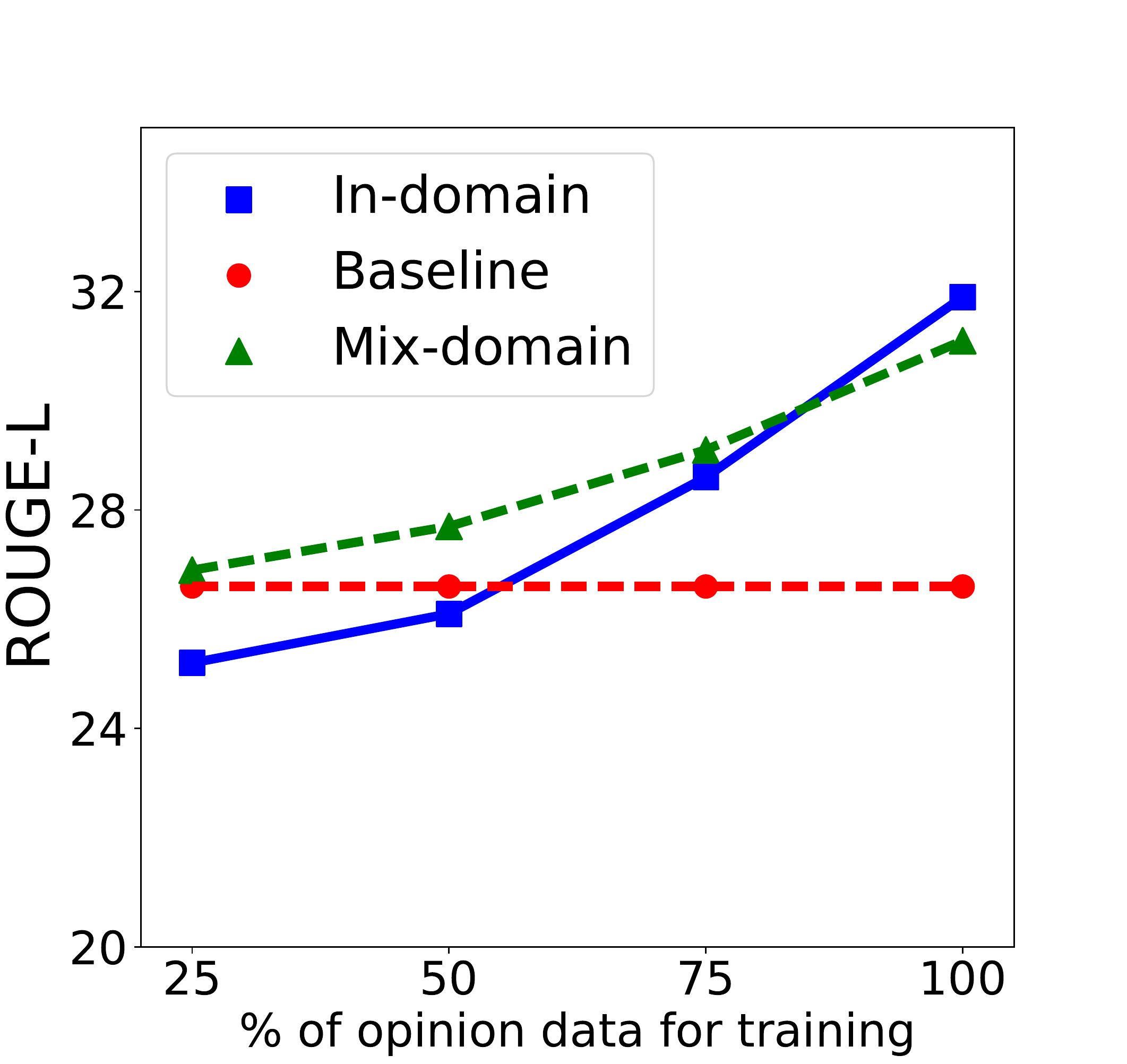}
}
 \caption{\fontsize{10}{12}\selectfont BLEU (left) and ROUGE-L (right) performance on In-domain and Mix-domain setup over different amount of training data. As the training data increases, In-domain outperforms Mix-domain training.}
\label{fig:eval_data}
\end{figure}

\noindent \textbf{Effect of Domain Adaptation.} 
Here we evaluate on domain adaptation, where \textsc{Opinion} is the target domain. From Figure~\ref{fig:eval_data}, we can see that when In-domain data is insufficient Mix-domain training yields better performance. As more In-domain training data becomes available, it outperforms Mix-domain training. Baseline for selecting the first sentence as summary is also displayed. 
Sample summaries in Figure~\ref{fig:samplesummary} also shows that \textsc{Out-of-Domain} training tends to generate summary in similar style to the source domain, while \textsc{Mix-Domain} training introduces the style of the target domain. 
{In our dataset, the first sentences of summaries for \textsc{Opinion} are usually in the form of \emph{[PERSON] reviews/criticizes/columns [EVENT]}, but the summaries for \textsc{News} usually start with event descriptions directly. Such style difference is reflected in \textsc{Out-of-Domain} and \textsc{Mix-Domain} too. }

\begin{figure}[ht]
	\fontsize{10}{12}\selectfont
	\setlength{\tabcolsep}{0.8mm}
	\begin{tabular}{|p{75mm}|}
	\hline
	\textbf{Human}: stephen holden reviews carnegie hall concert celebrating music of judy garland. singers include her daughter, lorna luft.  \\
	\textbf{Out-of-Domain}: article discusses possibility of carnegie hall in carnegie hall golf tournament.\\
	\textbf{Mix-Domain}: stephen holden reviews performance by jazz singer celebration by rainbow and garland at carnegie, part of tribute hall.\\
	\hline
	\end{tabular}
	
	\begin{tabular}{|p{75mm}|}
	\hline
	\textbf{Human}: janet maslin reviews john grisham book the king of torts .  \\
	\textbf{Out-of-Domain}: interview with john grisham of legal thriller is itself proof for john grisham 376 pages.\\
	\textbf{Mix-Domain}: janet maslin reviews book the king of torts by john grisham .\\
	\hline
	\end{tabular}
	
	\begin{tabular}{|p{75mm}|}
	\hline
	\textbf{Human}: anthony tommasini reviews 23d annual benefit concert of richard tucker music foundation , featuring members of metropolitan opera orchestra led by leonard slatkin .  \\
	\textbf{Out-of-Domain}: final choral society and richard tucker music foundation , on sunday night in [UNK] fisher hall , will even longer than substantive 22d gala last year .\\
	\textbf{Mix-Domain}: anthony tommasini reviews 23d annual benefit concert of benefit of richard tucker music.\\
	\hline
	\end{tabular}
	\caption{\fontsize{10}{12}\selectfont Sample summaries based on \textsc{Out-of-Domain} and \textsc{Mix-Domain} training on opinion articles.}
\label{fig:samplesummary}
\end{figure}

\begin{table}[t]
{
\fontsize{9}{11}\selectfont
\setlength{\tabcolsep}{0.5mm}
\begin{tabular}{|p{18mm}|c c c|c c c|c|}
\hline
\multirow{3}{*}{\textbf{}} & \multicolumn{6}{c|}{\textbf{Seen in Training} (\%)} & \multirow{3}{*}{\begin{minipage}{0.5in}\textbf{Unseen}\newline \centering{(\%)} \end{minipage}} \\ 
& \multicolumn{3}{c|}{\textit{In Input}} & \multicolumn{3}{c|}{\textit{Not In Input}} & \\
\cline{2-7}
& Gen & Mis & Total & Gen & Mis & Total & \\
\hline
\textit{Test on News} & & & & & & & \\
\textsc{In-Domain} & 33.7 & 40.9 & 74.6 & 2.6 & 19.3 & 21.9 & 3.5 \\
\hline\hline

\multicolumn{1}{|l|}{\textit{Test on Opinion}} & & & & & & & \\
\textsc{In-Domain} & 22.0 & 43.3 & 65.3 & 8.2  & 22.1 & 30.3 & \multirow{3}{*}{4.5} \\
\multicolumn{1}{|l|}{\textsc{Out-of-Domain}}  & 19.9 & 45.3 & 65.2 & 1.1 & 29.2 & 30.3 &  \\
\multicolumn{1}{|l|}{\textsc{Mix-Domain}}  & 18.6 & 46.6 & 65.2 & 6.3 & 23.9 & 30.2 &  \\
\hline
\end{tabular}

\caption{\fontsize{10}{12}\selectfont Comparison of generated (Gen) and missed (Mis) tokens for different training setups. We divide token in goldstandard summaries by 1) if it is seen in abstracts during training, and 2) if it is in the input text.}
\label{tab:breakdown}
}
\end{table}

We further classify the words in gold-standard summaries based on if they are seen in abstracts during training and then whether they are taken from the input text. We examine whether they are generated correctly. Full training set of opinion is used for in-domain and mix-domain training. Table~\ref{tab:breakdown} shows that among in-domain models, the model trained for news are superior at generating tokens mentioned in the input, compared to the model trained for opinion (33.7\% v.s. 22.0\%). Nonetheless, model trained for opinion is better at generating new words not in the input (8.2\% vs. 2.6\%). This is consistent with our observation that in opinion domain human editors favors new words different from the input.

\noindent \textbf{Further Analysis.} 
Here we study what information is transferable cross domains by investigating the attention weights assigned to the input text.

\noindent {\it What can be transferred.} We start with input words with highest attention weights when generating the summaries. Among these, we show the percentage over different word categories as in Table~\ref{tab:attention_dist}. 
For named entities, model trained on out-of-domain data pays more attention to PERSON and less attention to ORGANIZATION, while the in-domain trained model does reverse . This is consistent with the fact that opinion abstracts contains more PERSON and less ORGANIZATION than news abstracts (see Figure~\ref{fig:ne_subj_dist}). This suggests that the identification of summary-worthy named entities might be transferable from \textsc{news} to \textsc{opinion}. Similar effect is also observed for nouns and verbs, though less significant.

\noindent  {\it Attention change for domain adaptation.} We also examine the percentage of attention paid to summary-worthy words. For every output token we pick the input token with highest attention weight, and count the ones reused by human. For \textsc{In-Domain} test on \textsc{news}, on average 29.57\% of the output tokens have highest attention on summary-worthy words. For \textsc{Out-of-Domain} test on \textsc{opinion}, the number is 15.93\%; for \textsc{Mix-Domain}, it is 26.08\%. This shows the ability to focus on salient words is largely kept for \textsc{Mix-Domain} training.   
Additionally, as can be seen in Table~\ref{tab:attention_dist}, model trained on \textsc{Mix-Domain} puts more attention weights on PERSON (and all named entities) and nouns, but less attention on verbs and subjective words, compared with the model trained \textsc{Out-of-Domain}. This again aligns with our observation for the domain difference based on abstracts as in Figures~\ref{fig:pos_dist} and~\ref{fig:ne_subj_dist}.

\begin{table}[ht]
    \fontsize{9}{11}\selectfont
\setlength{\tabcolsep}{0.3mm}
    \begin{tabular}{|l|c|c|c|}
    \hline
    & \textsc{In-Domain} & \textsc{Out-of-Domain}  & \textsc{Mix-Domain} \\
    \textit{Src} $\rightarrow$ \textit{Trt}  & \textit{News} $\rightarrow$ \textit{News} &  \textit{News} $\rightarrow$ \textit{Opin} &  \multicolumn{1}{p{2cm}|}{\textit{News + Opin} $\rightarrow$ \textit{Opin}} \\
    \hline
    {PER} & 7.9\% & 8.7\% & \multicolumn{1}{c|}{15.1\% $\uparrow$} \\
    {ORG} & 10.9\% & 6.9\% & \multicolumn{1}{c|}{8.2\% $\uparrow$} \\
    {All NEs} & 26.7\% & 23.6\% & \multicolumn{1}{c|}{31.6\% $\uparrow$} \\
    \hline\hline
    {Noun} & 41.2\% & 36.2\% & \multicolumn{1}{c|}{43.3\% $\uparrow$} \\
    {Verb} & 10.3\% & 6.7\% & \multicolumn{1}{c|}{5.5\% $\downarrow$} \\
    \hline\hline
    {Positive} & 5.6\% & 5.1\% & \multicolumn{1}{c|}{4.5\% $\downarrow$} \\
    {Negative} & 2.5\% & 2.2\% & \multicolumn{1}{c|}{2.1\% $\downarrow$} \\
    \hline
    \end{tabular}
    \caption{\fontsize{10}{12}\selectfont
    Attention distribution on different word categories. We consider input words with highest attention weights when generating the summaries, and characterize them by Named Entity, POS tag, and Subjectivity.  The arrows shows the change with regard to \textsc{Out-of-Domain}. 
    }
    \label{tab:attention_dist}
\end{table}

\section{Related Work}
%\vspace{-1mm}
\label{sec:related}
Domain adaptation has been studied for a wide range of natural language processing tasks~\cite{blitzer2007biographies,florian-EtAl:2004:HLTNAACL,daumeiii:2007:ACLMain,Foster:2010:DIW:1870658.1870702}. However, little has been done for investigating summarization systems~\cite{sandu2010domain,wang-cardie:2013:ACL2013}. To the best of our knowledge, we are the first to study the adaptation of neural summarization models for new domain. 
Furthermore, Recent work in neural summarization mainly focuses on specfic extensions to improve system performance~\cite{rush-chopra-weston:2015:EMNLP,takase-EtAl:2016:EMNLP2016,gu-EtAl:2016:P16-1,nallapati2016abstractive,ranzato2015sequence}. It is unclear how to adapt the existing neural summarization systems to a new domain when the training data is limited or not available. This is a question we aim to address in this work.

%\vspace{-1mm}
\section{Conclusion}
%\vspace{-1mm}
\label{sec:conclusion}
We investigated domain adaptation for abstractive neural summarization. Experimental results showed that pre-training model with extractive summaries helps.
By analyzing the attention weight distribution over input tokens, we found the model was capable to select salient information even trained on out-of-domain data. 
This points to future direcions where domain adaptation techniques can be developed to allow a summarization system to learn content selection from out-of-domain data while acquiring language generating behavior with in-domain data.

\section*{Acknowledgments}
This work was supported in part by National Science Foundation Grant IIS-1566382 and a GPU gift from Nvidia. We thank three anonymous reviewers for their valuable suggestions on various aspects of this work.

\bibliography{domain,other}

\begin{thebibliography}{}
\expandafter\ifx\csname natexlab\endcsname\relax\def\natexlab#1{#1}\fi

\bibitem[{Bahdanau et~al.(2014)Bahdanau, Cho, and Bengio}]{bahdanau2014neural}
Dzmitry Bahdanau, Kyunghyun Cho, and Yoshua Bengio. 2014.
\newblock Neural machine translation by jointly learning to align and
  translate.
\newblock {\em arXiv preprint arXiv:1409.0473\/} .

\bibitem[{Blitzer et~al.(2007)Blitzer, Dredze, Pereira
  et~al.}]{blitzer2007biographies}
John Blitzer, Mark Dredze, Fernando Pereira, et~al. 2007.
\newblock Biographies, bollywood, boom-boxes and blenders: Domain adaptation
  for sentiment classification.
\newblock In {\em ACL\/}. volume~7, pages 440--447.

\bibitem[{Daume~III(2007)}]{daumeiii:2007:ACLMain}
Hal Daume~III. 2007.
\newblock \href{http://www.aclweb.org/anthology/P07-1033}{Frustratingly easy
  domain adaptation}.
\newblock In {\em Proceedings of the 45th Annual Meeting of the Association of
  Computational Linguistics\/}. Association for Computational Linguistics,
  Prague, Czech Republic, pages 256--263.
\newblock
  \href{http://www.aclweb.org/anthology/P07-1033}{http://www.aclweb.org/anthology/P07-1033}.

\bibitem[{Florian et~al.(2004)Florian, Hassan, Ittycheriah, Jing, Kambhatla,
  Luo, Nicolov, and Roukos}]{florian-EtAl:2004:HLTNAACL}
R~Florian, H~Hassan, A~Ittycheriah, H~Jing, N~Kambhatla, X~Luo, N~Nicolov, and
  S~Roukos. 2004.
\newblock A statistical model for multilingual entity detection and tracking.
\newblock In Daniel~Marcu Susan~Dumais and Salim Roukos, editors, {\em
  HLT-NAACL 2004: Main Proceedings\/}. Association for Computational
  Linguistics, Boston, Massachusetts, USA, pages 1--8.

\bibitem[{Foster et~al.(2010)Foster, Goutte, and
  Kuhn}]{Foster:2010:DIW:1870658.1870702}
George Foster, Cyril Goutte, and Roland Kuhn. 2010.
\newblock
  \href{http://dl.acm.org/citation.cfm?id=1870658.1870702}{Discriminative
  instance weighting for domain adaptation in statistical machine translation}.
\newblock In {\em Proceedings of the 2010 Conference on Empirical Methods in
  Natural Language Processing\/}. Association for Computational Linguistics,
  Stroudsburg, PA, USA, EMNLP '10, pages 451--459.
\newblock
  \href{http://dl.acm.org/citation.cfm?id=1870658.1870702}{http://dl.acm.org/citation.cfm?id=1870658.1870702}.

\bibitem[{Ganesan et~al.(2010)Ganesan, Zhai, and Han}]{ganesan2010opinosis}
Kavita Ganesan, ChengXiang Zhai, and Jiawei Han. 2010.
\newblock Opinosis: a graph-based approach to abstractive summarization of
  highly redundant opinions.
\newblock In {\em Proceedings of the 23rd international conference on
  computational linguistics\/}. Association for Computational Linguistics,
  pages 340--348.

\bibitem[{Gerani et~al.(2014)Gerani, Mehdad, Carenini, Ng, and
  Nejat}]{gerani2014abstractive}
Shima Gerani, Yashar Mehdad, Giuseppe Carenini, Raymond~T Ng, and Bita Nejat.
  2014.
\newblock Abstractive summarization of product reviews using discourse
  structure.
\newblock In {\em EMNLP\/}. pages 1602--1613.

\bibitem[{Gu et~al.(2016)Gu, Lu, Li, and Li}]{gu-EtAl:2016:P16-1}
Jiatao Gu, Zhengdong Lu, Hang Li, and Victor~O.K. Li. 2016.
\newblock \href{http://www.aclweb.org/anthology/P16-1154}{Incorporating copying
  mechanism in sequence-to-sequence learning}.
\newblock In {\em Proceedings of the 54th Annual Meeting of the Association for
  Computational Linguistics (Volume 1: Long Papers)\/}. Association for
  Computational Linguistics, Berlin, Germany, pages 1631--1640.
\newblock
  \href{http://www.aclweb.org/anthology/P16-1154}{http://www.aclweb.org/anthology/P16-1154}.

\bibitem[{Lin(2004)}]{lin2004rouge}
Chin-Yew Lin. 2004.
\newblock Rouge: A package for automatic evaluation of summaries.
\newblock In {\em Text summarization branches out: Proceedings of the ACL-04
  workshop\/}. Barcelona, Spain, volume~8.

\bibitem[{Nallapati et~al.(2016)Nallapati, Zhou, dos Santos, glar
  Gul{\c{c}}ehre, and Xiang}]{nallapati2016abstractive}
Ramesh Nallapati, Bowen Zhou, Cicero dos Santos, {\c{C}}a~glar Gul{\c{c}}ehre,
  and Bing Xiang. 2016.
\newblock Abstractive text summarization using sequence-to-sequence rnns and
  beyond.
\newblock {\em CoNLL 2016\/} page 280.

\bibitem[{Nenkova et~al.(2011)Nenkova, McKeown et~al.}]{nenkova2011automatic}
Ani Nenkova, Kathleen McKeown, et~al. 2011.
\newblock Automatic summarization.
\newblock {\em Foundations and Trends{\textregistered} in Information
  Retrieval\/} 5(2--3):103--233.

\bibitem[{Papineni et~al.(2002)Papineni, Roukos, Ward, and
  Zhu}]{papineni2002bleu}
Kishore Papineni, Salim Roukos, Todd Ward, and Wei-Jing Zhu. 2002.
\newblock Bleu: a method for automatic evaluation of machine translation.
\newblock In {\em Proceedings of the 40th annual meeting on association for
  computational linguistics\/}. Association for Computational Linguistics,
  pages 311--318.

\bibitem[{Pighin et~al.(2014)Pighin, Cornolti, Alfonseca, and
  Filippova}]{pighin-EtAl:2014:P14-1}
Daniele Pighin, Marco Cornolti, Enrique Alfonseca, and Katja Filippova. 2014.
\newblock \href{http://www.aclweb.org/anthology/P14-1084}{Modelling events
  through memory-based, open-ie patterns for abstractive summarization}.
\newblock In {\em Proceedings of the 52nd Annual Meeting of the Association for
  Computational Linguistics (Volume 1: Long Papers)\/}. Association for
  Computational Linguistics, Baltimore, Maryland, pages 892--901.
\newblock
  \href{http://www.aclweb.org/anthology/P14-1084}{http://www.aclweb.org/anthology/P14-1084}.

\bibitem[{Ramachandran et~al.(2016)Ramachandran, Liu, and
  Le}]{ramachandran2016unsupervised}
Prajit Ramachandran, Peter~J Liu, and Quoc~V Le. 2016.
\newblock Unsupervised pretraining for sequence to sequence learning.
\newblock {\em arXiv preprint arXiv:1611.02683\/} .

\bibitem[{Ranzato et~al.(2015)Ranzato, Chopra, Auli, and
  Zaremba}]{ranzato2015sequence}
Marc'Aurelio Ranzato, Sumit Chopra, Michael Auli, and Wojciech Zaremba. 2015.
\newblock Sequence level training with recurrent neural networks.
\newblock {\em arXiv preprint arXiv:1511.06732\/} .

\bibitem[{Rush et~al.(2015)Rush, Chopra, and
  Weston}]{rush-chopra-weston:2015:EMNLP}
Alexander~M. Rush, Sumit Chopra, and Jason Weston. 2015.
\newblock \href{http://aclweb.org/anthology/D15-1044}{A neural attention model
  for abstractive sentence summarization}.
\newblock In {\em Proceedings of the 2015 Conference on Empirical Methods in
  Natural Language Processing\/}. Association for Computational Linguistics,
  Lisbon, Portugal, pages 379--389.
\newblock
  \href{http://aclweb.org/anthology/D15-1044}{http://aclweb.org/anthology/D15-1044}.

\bibitem[{Sandhaus(2008)}]{sandhausnew}
Evan Sandhaus. 2008.
\newblock The new york times annotated corpus, 2008.
\newblock {\em Linguistic Data Consortium, PA\/} .

\bibitem[{Sandu et~al.(2010)Sandu, Carenini, Murray, and Ng}]{sandu2010domain}
Oana Sandu, Giuseppe Carenini, Gabriel Murray, and Raymond Ng. 2010.
\newblock Domain adaptation to summarize human conversations.
\newblock In {\em Proceedings of the 2010 Workshop on Domain Adaptation for
  Natural Language Processing\/}. Association for Computational Linguistics,
  pages 16--22.

\bibitem[{See et~al.(2017)See, Liu, and Manning}]{see2017get}
Abigail See, Peter~J Liu, and Christopher~D Manning. 2017.
\newblock Get to the point: Summarization with pointer-generator networks.
\newblock {\em arXiv preprint arXiv:1704.04368\/} .

\bibitem[{Takase et~al.(2016)Takase, Suzuki, Okazaki, Hirao, and
  Nagata}]{takase-EtAl:2016:EMNLP2016}
Sho Takase, Jun Suzuki, Naoaki Okazaki, Tsutomu Hirao, and Masaaki Nagata.
  2016.
\newblock \href{https://aclweb.org/anthology/D16-1112}{Neural headline
  generation on abstract meaning representation}.
\newblock In {\em Proceedings of the 2016 Conference on Empirical Methods in
  Natural Language Processing\/}. Association for Computational Linguistics,
  Austin, Texas, pages 1054--1059.
\newblock
  \href{https://aclweb.org/anthology/D16-1112}{https://aclweb.org/anthology/D16-1112}.

\bibitem[{Vinyals et~al.(2015)Vinyals, Fortunato, and
  Jaitly}]{vinyals2015pointer}
Oriol Vinyals, Meire Fortunato, and Navdeep Jaitly. 2015.
\newblock Pointer networks.
\newblock In {\em Advances in Neural Information Processing Systems\/}. pages
  2692--2700.

\bibitem[{Wang and Cardie(2013)}]{wang-cardie:2013:ACL2013}
Lu~Wang and Claire Cardie. 2013.
\newblock \href{http://www.aclweb.org/anthology/P13-1137}{Domain-independent
  abstract generation for focused meeting summarization}.
\newblock In {\em Proceedings of the 51st Annual Meeting of the Association for
  Computational Linguistics (Volume 1: Long Papers)\/}. Association for
  Computational Linguistics, Sofia, Bulgaria, pages 1395--1405.
\newblock
  \href{http://www.aclweb.org/anthology/P13-1137}{http://www.aclweb.org/anthology/P13-1137}.

\bibitem[{Wang and Ling(2016)}]{wang-ling:2016:N16-1}
Lu~Wang and Wang Ling. 2016.
\newblock \href{http://www.aclweb.org/anthology/N16-1007}{Neural network-based
  abstract generation for opinions and arguments}.
\newblock In {\em Proceedings of the 2016 Conference of the North American
  Chapter of the Association for Computational Linguistics: Human Language
  Technologies\/}. Association for Computational Linguistics, San Diego,
  California, pages 47--57.
\newblock
  \href{http://www.aclweb.org/anthology/N16-1007}{http://www.aclweb.org/anthology/N16-1007}.

\bibitem[{Wilson et~al.(2005)Wilson, Wiebe, and
  Hoffmann}]{wilson2005recognizing}
Theresa Wilson, Janyce Wiebe, and Paul Hoffmann. 2005.
\newblock Recognizing contextual polarity in phrase-level sentiment analysis.
\newblock In {\em Proceedings of the conference on human language technology
  and empirical methods in natural language processing\/}. Association for
  Computational Linguistics, pages 347--354.

\end{thebibliography}
\bibliographystyle{acl_natbib}

\end{document}